# Visual Passwords Using Automatic Lip Reading

Ahmad B. A. Hassanat[*]

*IT Department, Mu'tah University, Mu'tah – Karak, Jordan, 61710.*

*hasanat@mutah.edu.jo*

**Abstract**

This paper presents a visual passwords system to increase security. The system depends mainly on recognizing the speaker using the visual speech signal alone. The proposed scheme works in two stages: setting the visual password stage and the verification stage.

At the setting stage the visual passwords system request the user to utter a selected password, a video recording of the user face is captured, and processed by a special words-based VSR system which extracts a sequence of feature vectors. In the verification stage, the same procedure is executed, the features will be sent to be compared with the stored visual password.

The proposed scheme has been evaluated using a video database of 20 different speakers (10 females and 10 males), and 15 more males in another video database with different experiment sets. The evaluation has proved the system feasibility, with average error rate in the range of 7.63% to 20.51% at the worst tested scenario, and therefore, has potential to be a practical approach with the support of other conventional authentication methods such as the use of usernames and passwords.

1. Introduction

Identity theft is one of the fastest growing crimes in the world, it is estimated that 7.2 millions become identity theft and/or fraud victims in the last 13 years in the United States alone [1]. Electronic fund transfer-related identity theft was the most frequently reported type of identity theft bank fraud during calendar year 2008. The same reference stated that the annual fraud amount paid in the year 2008 was more than $1.8 billion in the USA alone. The methods of initial contact of those fraud complaints include 52% by email, 11% by Internet websites and only 7% of those fraud complaints were through the phone as the initial point of contact. All these numbers came from claiming others identities using what they know, because what they know can be known by impostors using many methods. Current conventional authentication schemes are based only on "what you know?" such as usernames, passwords, PIN numbers, etc., all consisting of numbers, alphabetic letters and special characters. These approaches are vulnerable to a wide range of attacks including brute-force methods and phishing.

Biometric-based authentication provides a stronger alternative since it is based on "who you are?" Such schemes include the use of fingerprint, iris and face authentication. The Visual passwords based authentication is a behavioral biometrics depends on recognizing the speaker using the visual signal alone. Unlike most biometrics the proposed scheme does not need special equipments (only a camera), and it is difficult to be deceived using a face image in front of the camera.

------------------------------------------------------------------------
* Corresponding author.
E-mail address: hasanat@mutah.edu.jo.

Sometimes people whisper to each other to prevent others from hearing them, and sometimes they do not even generate any Sound, and the other side has to understand what they meant using their faces visibility, visual cues and lip-reading. The visual passwords scheme tries to mimic this behavior by allowing the client to whisper the password to a secured system, without having the need to produce the audio signal, in order to increase security in such computerized and human-computer interaction systems.

The visual password scheme processes visual signal feature, obtained from a video of the speaker while uttering a special sequence of words. This includes measuring some aspects of the lip movements and appearance, and finding a match between the spoken word(s) and the assigned password. Accordingly, the scheme grants or declines access to the secured system. A visual password is the visual side of a spoken word without incorporating the audio signal, so clients can provide their passwords by moving their lips without revealing the sound of the uttered word. This is beneficial, particularly to prevent others from hearing the password, and also work fine in noisy environment, where the audio signal become useless. Basically, the visual passwords scheme is a visual only speaker authentication.

On the other hand we can't neglect the richer and more important speech information embedded in the audio signal (which is out of the scope of this study), as speech production is bimodal in nature, which is produced by the vibration of the vocal cord and the configuration of the vocal tract, including the nasal cavity, the tongue, teeth, velum, and lips [2]. This paper focuses on the visual side of speech only, particularly, when the audio signal is not generated or was noisy.

The rest of this paper is devoted to review some of the related work, particularly our previous work [3] [4] & [5] on which the visual passwords scheme is partially based, in addition to presenting the proposed scheme, and discussing the related experiments and their results.

## 2. Related Work

Most of the work done on speaker recognition was based on the audio signal alone, or on integrating the audio and visual signal, because the visual signal contains less information than the audio signal. Nonetheless facial visual cues can improve speech perception because the visual signal is correlated to the audio signal and contains complementary information to it [6] [7] [8] [9] & [10].

The first speaker recognition system using integrated acoustic and visual speech was proposed by Chibelushi and co-workers. For the acoustic features they used the Perceptual Linear Predictive (PLP) cepstral features to represent the vocal speaker characteristics. For the visual features they used a kinetic model of lip motion for extracting the characteristics of the changes in lip shape during speech. An artificial neural network (ANN) model was employed for the classification step. According to Chibelushi and his colleagues, the identification error rates were reduced to 1.25% using the integrated audio-visual recognition system [11].

Civanlar and Chen [12] used the mouth height and width as visual features, and the LPC coefficients as the audio features. Those features were combined to form the feature vector, and weighted depending on the acoustic noise. Dynamic time warping was used for recognition using a predefined threshold. The reported error rate for voice-only speaker verification was about 40% for noisy speech signal, compared to 5% for clean speech signal. The error rate for noisy signal was reduced to 10% using joint audiovisual verification. But for clean signal, incorporating the visual signal did not reduce the error significantly.

Chaudhari and co-workers [13] proposed an audio-visual speaker identification and verification. The audio were represented by 23 Mel-frequency cepstral coefficients (MFCC) and the visual parameters by 24 Discrete Cosine Transform (DCT) coefficients of the mouth region. Gaussian Mixture Models (GMMs) were used to model the speakers. The system was evaluated on the IBM ViaVoice database, the accuracy of their system using audio only was 95.5% increased to 99.5% after integrating the video signal.

Faraj and Bigun [14] proposed an identity authentication technique by a synergetic use of lip-motion and speech, the lip-motion is defined as the distribution of apparent velocities in the movement of brightness patterns in an image. The audio features were extracted using the Mel-frequency cepstral coefficients (MFCC). Both audio and video information were fused to be recognized using GMMs. They evaluated their system using the XM2VTS database claiming a 98% recognition rate. Other researchers who worked on speaker recognition using AV speech information, include but not limited to [15] & [16].

Relatively very little work has been reported in the literature concerning the speaker recognition problem using the visual signal alone. This includes the work of Shiell and co-workers [17], who proposed a visual only system for speaker identification using Viola and Jones's method for face detection [18], active appearance model (AAM) [19] and DCT for visual features extraction. Hidden Markov models (HMM) was employed for classification. The system was evaluated using "VALID" database [20], their reported speaker identification rate was 59.3%

Cetingul and co-workers [21] proposed a quasi-automatic system to extract and analyze robust lip-motion features, the initial lip motion was represented using 2 approaches, 2D-DCT of the mouth region, and the other is tracking the lip's boundaries over the video frames. An HMM-based identification system is used to compare the results of both approaches. The system was evaluated on the audio-visual database MVGL-AVD [22]. Average error rates (AER) for the first approach were in the range of (8.4-8.9%) depending on the size of the grid, and (8.4-11.2%) depending on the size of the shape feature vector, in both cases the best performance obtained with the largest size. A more in-depth review can be found in [23].

Shafait [24] proposed an algorithm to differentiate between a real person and a photograph using lip motion analysis. He recommend the use of 2D-DCT for feature extraction, the classifiers were either GMM or k-nearest neighbour (KNN), he evaluated his system using the "BioID" face database, reporting two recognition rates, 86% using the KNN, and 83% using GMM.

Our previous work on automatic lip reading termed as visual words (VWords) [3] & [5] fits also in this category; this system provides a full automated solution for the visual (only) speech recognition (VSR) problem. The system (see Figure 1) consists of three major stages: detecting/localizing human faces [4], lip localization [25] and lip reading.

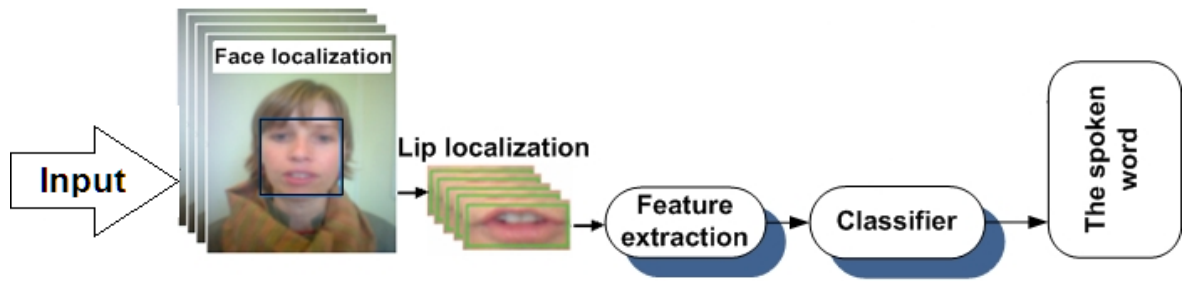

Fig. 1. Automatic lip-reading system (Vwords).

The first stages are discussed thoroughly in [4] & [25], the lip reading stage consists of 2 steps, feature extraction and word recognition. The following is the list of features, which are extracted from the sequences of the mouth area (region of interest (ROI)) during the uttering of a word (see Figure 3):

1. The height (H) and width (W) of the mouth, i.e. ROI height and width (geometric-based features).

2. The mutual information (M) between consecutive frames ROI in the discrete wavelet transform (DWT) domain (image-transformed-based features based on temporal information). The mutual information M between two random variables X: DWT (current mouth) and Y: DWT (previous mouth) is defined by:

$$M(X;Y) = \sum_x \sum_y p(x,y) \log\left(\frac{p(x,y)}{p(x)p(y)}\right) \quad (1)$$

3. The image quality value (Q) of the current ROI with reference to its predecessor measured in the DWT domain (image-transformed-based features based on temporal information). Image quality models image distortion as a combination of loss of correlation, luminance distortion, and contrast distortion. The quality measure Q is given by:

$$Q = \left(\frac{4\sigma_{xy}\overline{xy}}{(\sigma_x^2 + \sigma_y^2)[(\overline{x})^2 + (\overline{y})^2]}\right) \quad (2)$$

4. The ratio of vertical to horizontal features (R) taken from DWT of ROI (image-transformed-based features based on temporal information). The ratio (R) of the vertical features obtained from wavelet sub-band HL to the number of the horizontal ones gained from the LH is given by:

$$R = \frac{V}{H} \quad (3)$$

Where V = number of vertical features, and H = number of horizontal features. It is well established that redundant data are located around the medians of the non-LL sub-bands, and detailed data are located further, accordingly, by substituting V and H

in equation 3, we get:

$$R = \frac{\sum_x \sum_y \begin{cases} 0 & (HL_{median} - \sigma_{HL}) \leq HL(x,y) \leq (HL_{median} + \sigma_{HL}) \\ 1 & otherwise \end{cases}}{\sum_x \sum_y \begin{cases} 0 & (LH_{median} - \sigma_{LH}) \leq LH(x,y) \leq (LH_{median} + \sigma_{HL}) \\ 1 & otherwise \end{cases}} \quad (4)$$

5. The ratio of vertical edges to horizontal edges (ER) of ROI (image-transformed-based features). ER is obtained by using the Sobel edge detector [26] which is given by:

$$ER = \frac{\sum_x^W \sum_y^H \sum_{i=-1}^{1} \sum_{j=-1}^{1} |ROI(x+i,y+j)(S_v(i+1,j+1))|}{\sum_x^W \sum_y^H \sum_{i=-1}^{1} \sum_{j=-1}^{1} |ROI(x+i,y+j)(S_h(i+1,j+1))|} \quad (5)$$

6. The amount of red color (RC) in ROI as an indicator of the appearance of the tongue (image-appearance-based features). RC is the ratio of the red color to the size of ROI, this ratio can be calculated using the following equation:

$$RC = \frac{\sum_{x=1}^{W} \sum_{y=1}^{H} red(ROI(x,y))}{(W)(H)} \quad (6)$$

7. The amount of visible teeth (T) in the ROI (image-appearance-based features). Detecting teeth in the ROI is a visual cue for uttering some phonemes and enriches the visual words signatures; the major characteristic that distinguishes teeth from other parts of the ROI is the low saturation and high intensity values [27]. By converting the pixels values of ROI to 1976 CIELAB color space (L*, a*, b*) and 1976 CIELUV color space (L*, u*, v*), the teeth pixel has a lower a* and u* value than other lip pixels [28]. A teeth pixel can be defined by:

$$t = \begin{cases} 1 & a^* \leq (\mu_a - \sigma_a) \\ 1 & u^* \leq (\mu_u - \sigma_u) \\ 0 & otherwise \end{cases} \quad (7)$$

The appearance of the teeth can be defined by the number of teeth pixels in ROI. Therefore, the amount of teeth in ROI is given by:

$$T = \sum_{x=1}^{W} \sum_{y=1}^{H} t(x,y) \quad (8)$$

The previous features were proven to be effective for automatic lip reading, so they will be used for the experiments conducted for this paper. In addition, this paper proposes adding an extra appearance-based feature to the features set as an attempt to increase accuracy; this feature comes from the Chi-square test, which is applied on the histogram of the current mouth image (observed values) and the histogram of the first mouth image (expected values). Ideally, the first image of the mouth (for each word's frame series) represents silence. Chi-square test reveals the amount of similarity between two variables, similarity between the current frame and the silence frame is affected by the changes of the mouth shape that occurs when uttering different phonemes, see figure (2). Chi-square test could be computed using equation 9.

$$\chi^2 = \sum_{i=0}^{255} \frac{(O_i - E_i)^2}{E_i} \quad (9)$$

Where $O_i$ is the frequency of intensity $i$ in the observed mouth frame, and $E_i$ is the frequency of intensity $i$ in the first mouth frame. If one or more of the expected values were zeros, those values are not computed.

The previous equation yields large numbers, and the histogram values are affected by the size of the mouth, so it is better to use the probability of each color rather than the frequency using the following equation.

| 1 | 2 | 3 | 4 | 5 | 6 | 7 | 8 | 9 | 10 | 11 |
|---|---|---|---|---|---|---|---|---|---|---|
| 0.00 | 0.26 | 0.16 | 0.30 | 0.30 | 0.32 | 0.33 | 0.30 | 0.24 | 0.21 | 0.26 |
| 12 | 13 | 14 | 15 | 16 | 17 | 18 | 19 | 20 | 21 | 22 |
| 0.26 | 0.26 | 0.26 | 0.28 | 0.31 | 0.27 | 0.20 | 0.28 | 0.31 | 0.33 | 0.72 |
| 23 | 24 | 25 | 26 | 27 | 28 | 29 | 30 | 31 | | |
| 0.60 | 0.50 | 0.70 | 0.84 | 0.88 | 1.18 | 1.04 | 0.36 | 0.57 | | |

Fig. 2. The appearance of the mouth while uttering the word "zero", in addition to the values of the chi-square test below each frame.

$$\chi^2 = \sum_{i=0}^{255} \frac{\left(\dfrac{O_i}{w_1.h_1} - \dfrac{E_i}{w_2.h_2}\right)^2}{\dfrac{E_i}{w_2.h_2}} \qquad (10)$$

Where $w_1$ and $h_1$ are width and height of observed mouth, $w_2$ and $h_2$ are width and height of the first mouth.

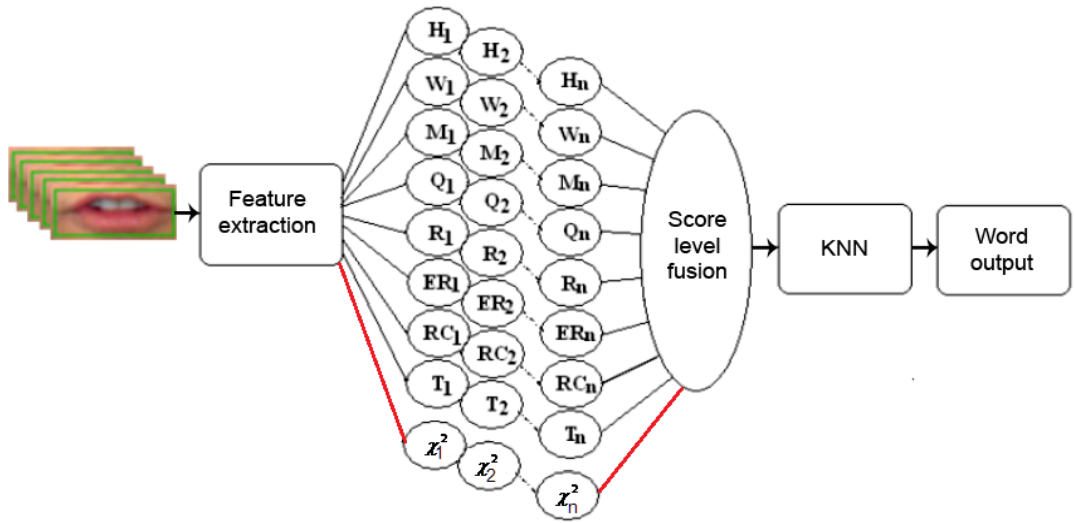

Fig. 3. Visual words feature extraction and recognition method

For the recognition step the KNN method is employed for two reasons. Firstly, the length of each word is different from person to another, and from time to time even those uttered by the same person, so interpolation is used to stretch the smaller signal to the larger one when comparing. Secondly, KNN is the simplest classifier and it is efficient when the training data is very small, and this is the case in this paper. Moreover, "the no free lunch theorem of optimization", suggests that there is no general-purpose universal optimization strategy [29], therefore, there is no need to try other classifiers, at least for the purpose of this paper.

Unlike the state-of-the-art VSR approaches, which depend mainly on recognizing a visual speech unit called "Viseme" (the visual part of a phoneme), we developed a holistic approach to VSR that is based on visually recognizing words rather than their parts. This is done by analyzing several visual signals (see Figure 3) that can be associated with each word. Each of these signals represents the variation of a specific visual feature over the time, taken when the word is spoken.

It turned out that the holistic approach preserves the speaker identity better than the visemic one. And we found that the VSR problem in general is a speaker dependent problem, and become more dependent when using the signature of the whole word as the individual differences are preserved. This is also fostered by the use of appearance based features (such as RC, T and X), which also preserves the different appearance of the ROI, which different from one to another [3]. Therefore we used the visual words approach as the base for the visual passwords scheme.

## 3. The visual passwords scheme

The proposed scheme works in two stages: setting the visual password stage and the verification stage. At the setting stage the visual passwords system requests the user to utter a selected password consisting of a number of words, a video recording of the user face is captured, and processed by the discussed VSR system, which extracts a sequence of feature vectors, one for each word in the password. Each feature vector is represented by a matrix, each row of which represents a frame of the video, associated with the given word, representing some lip region measurements including width, height, mutual information between successive frames, image quality, horizontal/vertical features ratio, teeth, tongue appearance and chi-square test. The sequence of extracted matrices (feature vectors) is stored on the authentication server.

In the verification stage, the same procedure is executed to extract the sequence of feature vectors, for the set of words uttered by the claimant when prompted, which would be sent in an encrypted form by the client to the authentication server. On receipt the server decrypt and compare with the stored visual password feature vectors.

The main task of the "Visual Passwords" includes training a secure system on some examples of a particular Vword of a particular speaker, giving only their visual word signal; how is it possible to grant him/her access? And giving an unknown Vword how is it possible to decline him/her access? Those are the major research questions.

The results from our previous work [3] show that VSR is a speaker-dependent problem, since the speaker-independent word recognition rate is about 32% on average, i.e. the word error rate is about 68% on average, which means that the same word spoken by two different speakers is more likely (68%) to produce two different signatures. This difference varies between people, and depends mainly on the individual differences between them, such as the way they speak (speech style and behavior), and the different appearance of the lips and mouth region.

These individual differences form something like "visual speechprint", which can be beneficial in applications such as the Visual passwords, which is the process of granting/declining access to a speaker who is speaking on the basis of their (unique to some extent) individual information embedded in their speech signal.

Visual passwords (VPs), which can be generated using VWords, can be used as a backup for the normal string passwords, or as an alternative to face verification, as the latter can be faked by putting a fake image in front of a camera. For more security, visual passwords can be obtained for a system by asking the user to say a specific word several times. This word can be chosen by the user or provided by the system. Even if someone hacks into the string password, he will not be able to produce a visual word in the same manner that the client did, because of individual differences in mouth appearance, and the different way people speak and speak visually. Thus, the security of that system should be increased.

The VPs problem can be thought of as a speaker verification problem, where the examples of only one speaker (client) is known previously, and the computer has to authenticate the speaker of a VWord approximately equal to one of those in the training set (system database). A specific threshold needs to be defined to approximate the test words, i.e. if the distance between the test word and the visual password is less than a threshold, then access to the system is granted, otherwise access to the system is declined.

In other words, if the distance (D) of the nearest neighbor (NN) is less than the threshold (T), then the claimer is granted access to the system, otherwise, the claimer is given several tries to log in, and if he/she fails after n tries, he/she is considered an impostor, and the system will then block him/her (see Figure 4). In this work we used Euclidian distance and linear interpolation to compensate for the variation in the words lengths.

To test the visual passwords scheme in this paper, 10 females and 10 males were randomly selected from an in-house video database, which is described in [3]. This database consists of 26 participants (10 females and 16 males) of different races and nationalities (Africans, Europeans, Asians and Middle-Eastern volunteers) an equal number of males and females is chosen to avoid gender bias, particularly, if we know that facial hair may affect accuracy of lip segmentation and feature extraction.

Each participant recorded 2 videos (sessions 1 and 2) at different times. About half of the males have facial hair. The videos were recorded inside a normal room, each person in each recorded video utters non-contiguous 30 different words five times, which are numbers (from 0-9), short look-alike words (*knife, light, kit, night, fight*) and (*fold, sold, hold, bold, cold)*, long words: (*appreciate, university, determine, situation, practical*) and five security related words (*bomb, kill, run, gun, fire*).

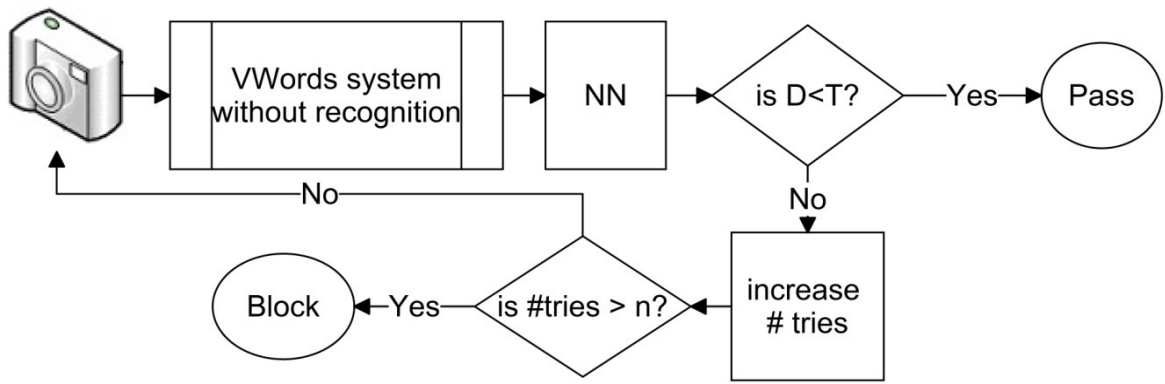

Fig. 4. The proposed visual passwords system.

**4. Experimental results**

There are 2 scenarios to be considered, the first being if the impostor knows the password and wants to produce it visually to access the system. And the other is when the impostor does not know the password and tries to say any word to access the system.

Assuming that the client is one of the selected subjects and the rest are 19 impostors. The word "zero" spoken by subject 1 was assumed to be the visual password. To define the suitable threshold, the visual password "*zero*" (spoken by subject 1 from session 2, repeated 5 times) was used as a training set, while the test set was the "zeros" spoken by the rest of the subjects from session 1. After defining the threshold, the system was evaluated using a test set of the same subject, but this time from session 2 instead of session 1, and the training set was the "zeros" spoken by the client (subject 1), but this time from session 1 instead of session 2, thus training the threshold in one session, and testing the method in the other session using the best threshold, see figure (5). This is done for each subject, as each subject is assumed each time to be the client and the rest are impostors.

| | Thresholding | | |
|---|---|---|---|
| | Speaker | Session | VWord |
| Training set | Subject 1 (Client) | 2 | Zero |
| Test set | All including the client | 1 | Zero |
| | Evaluation using threshold | | |
| | Speaker | Session | VWord |
| Training set | Subject 1 (Client) | 1 | Zero |
| Test set | All including the client | 2 | Zero |

Fig. 5. The visual passwords' evaluation protocol (scenario 1).

To find the best threshold which minimizes errors in the system, the visual password experiment was repeated 90 times, starting with threshold = 1, then increasing it by 0.1. There are 2 types of errors in such experiments, the false rejection error (FRE) and the false acceptance error (FAE). The FRE occurs when the system rejects the client considering him/her as an impostor, and the FAE occurs when the system accepts an impostor as a client. Both the false rejection ratio (FRR) and the false acceptance ratio (FAR) were recorded each time the threshold was increased (see Figure 6). The best threshold is the one where the total of FAR and FRR is minimized.

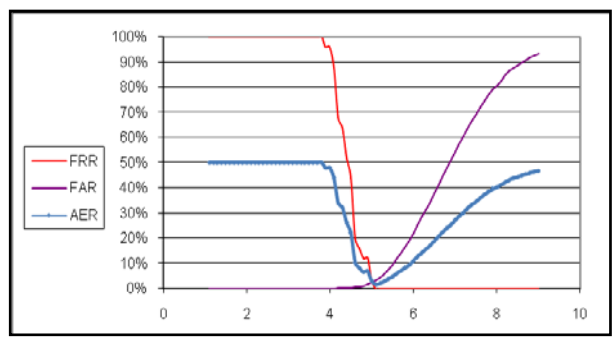

Fig. 6. Example of threshold selection for the visual password "fourfive", which was uttered by subject "Male02" in experiment DoubleVPunknown, the best AER (0.21%) achieved when threshold was 5.1.

We designed one experiment for each scenario, for the first scenario, assuming that the impostors know the password; the test set contains only the word "zero" spoken by 20 subjects. And the training set contains the word "zero" spoken by a client from the other session. The VP is a single word; we called this experiment (SingleVPknown).

In the second scenario, assuming that the impostors do not know the password, the test set contains a non-zero VP, which is the word "one" in this paper, which was spoken by 19 subjects (the impostors trying to log into the system), in addition to 5 examples of the word "zero", which was spoken by the client. The training set contains the word "zero" spoken by the client from the other session. This experiment is called (SingleVPunknown).

In both experiments, the number of the tested words is 20 (subjects) x 5 (examples each word) = 100. And the number of the trained words is 5. both experiments were repeated 20 times (one for each speaker), each time the speaker being evaluated being considered as a client, by training his/her Vwords, and the other 19 speakers being considered as impostors, trying to get into the system by providing their Vwords. The results are shown in Table 1 and 2.

It can be noticed from Table 1 and 2 the improvement in the performance of the visual passwords, particularly the reduction of the false acceptance error FAR when the VP is unknown to the impostors. Hence the new VP (that was used by the impostors to gain access to the system) has a different visual speechprint, so the probability to block them increased. In other hand, the FRR remained the same, because the word "zero" spoken by the client has not been changed.

Table 1. Visual passwords results, using "zero=>zero" in "SingleVPknown", and "one=>zero" in "SingleVPunknown" as the visual passwords.

| Subject | Threshold | SingleVPknown | | | SingleVPunknown | | |
|---|---|---|---|---|---|---|---|
| | | FRR | FAR | AER | FRR | FAR | AER |
| Female01 | 2 | 20% | 7.53% | 13.77% | 20% | 1.01% | 10.51% |
| Female02 | 3.7 | 0% | 66.67% | 33.34% | 0% | 25.25% | 12.63% |
| Female03 | 1.8 | 0% | 0.42% | 0.21% | 0% | 0.42% | 0.21% |
| Female04 | 2.9 | 20% | 27.96% | 23.98% | 20% | 2.02% | 11.01% |
| Female05 | 4.7 | 0% | 96.77% | 48.39% | 0% | 91.92% | 45.96% |
| Female06 | 1.4 | 0% | 4.23% | 2.12% | 0% | 0.00% | 0.00% |
| Female07 | 4.4 | 60% | 76.34% | 68.17% | 60% | 49.49% | 54.75% |
| Female08 | 2.7 | 0% | 61.29% | 30.65% | 0% | 33.33% | 16.67% |
| Female09 | 2.6 | 20% | 12.90% | 16.45% | 20% | 0.0303 | 0.11515 |
| Female10 | 1.7 | 0% | 8.60% | 4.30% | 0% | 0.00% | 0.00% |
| | Average | 12% | 36.27% | 24.14% | 12.00% | 20.65% | 16.32% |
| Male01 | 2.7 | 40% | 37.63% | 38.82% | 40% | 24.24% | 32.12% |
| Male02 | 2.7 | 80% | 24.73% | 52.37% | 80% | 26.26% | 53.13% |
| Male03 | 2.6 | 0% | 16.13% | 8.07% | 0% | 3.03% | 1.52% |
| Male04 | 2.1 | 0% | 21.51% | 10.76% | 0% | 12.12% | 6.06% |
| Male05 | 2.2 | 20% | 22.58% | 21.29% | 20% | 9.09% | 14.55% |
| Male06 | 2.5 | 0% | 34.41% | 17.21% | 0% | 17.82% | 8.91% |
| Male07 | 2.7 | 0% | 27.96% | 13.98% | 0% | 0.00% | 0.00% |
| Male08 | 2.6 | 40% | 8.60% | 24.30% | 40% | 0.00% | 20.00% |
| Male09 | 2 | 0% | 3.23% | 1.62% | 0% | 0.00% | 0.00% |
| Male10 | 1.6 | 0% | 1.06% | 0.53% | 0% | 0.00% | 0.00% |
| | Average | 18% | 19.78% | 18.89% | 18.0% | 9.3% | 13.6% |
| | **Overall average** | **15%** | **28.03%** | **21.51%** | **15.0%** | **15.0%** | **15.0%** |

The performance of the proposed visual passwords system can be further improved by considering more than one visual word in the training set. This will increase the length of the visual password, and provide a stronger signal (complex password), which reduces the probability of being hacked. Using the same protocol as for the visual passwords experiments and the same 20 subjects: for each speaker (client) all the possible combinations of the words "four" and "five" from session 1 have been concatenated to create a new signal, each word being repeated 5 times, so the number of concatenated words is 25 new signals (double Vwords), which is the size of the training set for each speaker, when evaluating each speaker as a client and the other 19 speakers as impostors.

Hence the data at hand is relatively small, it is useful to use the word concatenation, also the way of saying the word "A" might be different from the way of saying it next time, so mixing all these different ways of saying the word "A" with all these different ways of saying the word "B", for instance, enriches the training set by covering more possibilities, in addition to a better validation when using an affluent test set.

For scenario 1, the test set contains all the possible combinations of the words ("four" and "five") spoken by the rest of the speakers, in addition to all the possible combinations of the words ("four" and "five") spoken by the client in session 2. This experiment is called (DoubleVPknown).

For scenario 2, The test set contains all the possible combinations of 2 different words (the words "six" and "seven") spoken by the rest of the speakers in addition to all the possible combinations of the words ("four" and "five") spoken by the client in session 2. This experiment is called (DoubleVPunknown). The same previous evaluation protocol was used to find the best threshold (for each speaker), and to evaluate the system. Figure 7 depicts the performance of the system during the thresholding stage. In both experiments, the number of the tested examples is the square of (19 (subjects) x 5 (examples each word)) = 9025 double VPs, in addition to 25 examples from the client. And the number of the trained examples is 25 double VPs.

Both experiments were repeated 20 times (one for each speaker), each time the speaker being evaluated being considered as a client, and the other 19 speakers being considered as impostors. The results are shown in Tables 3 and 4.

Table 2. Visual passwords results, using "zero=>zero" in "SingleVPknown", and "one=>zero" in "SingleVPunknown" as the visual passwords, adding the Chi-square test feature.

| Subject | Threshold | SingleVPknown | | | SingleVPunknown | | |
|---|---|---|---|---|---|---|---|
| | | FRR | FAR | AER | FRR | FAR | AER |
| Female01 | 2.8 | 20% | 7.00% | 13.50% | 20% | 0% | 10.00% |
| Female02 | 4.0 | 0% | 59.00% | 29.50% | 0% | 22% | 11.05% |
| Female03 | 1.9 | 0% | 0.38% | 0.19% | 0% | 0% | 0.10% |
| Female04 | 3.7 | 20% | 22.75% | 21.38% | 20% | 3% | 11.50% |
| Female05 | 5.7 | 20% | 88.60% | 54.30% | 20% | 83% | 51.50% |
| Female06 | 1.7 | 0% | 6.00% | 3.00% | 0% | 3% | 1.50% |
| Female07 | 4.6 | 40% | 71.60% | 55.80% | 40% | 40% | 40.13% |
| Female08 | 3.6 | 20% | 56.80% | 38.40% | 20% | 24% | 22.20% |
| Female09 | 3.1 | 0% | 13.68% | 6.84% | 0% | 2% | 1.00% |
| Female10 | 2.0 | 0% | 5.50% | 2.75% | 0% | 0% | 0.00% |
| | Average | 12% | 33.13% | 22.57% | 12.00% | 17.80% | 14.90% |
| Male01 | 3.3 | 40% | 31.60% | 35.80% | 40.00% | 18.77% | 29.39% |
| Male02 | 3.6 | 60% | 26.80% | 43.40% | 60.00% | 27.29% | 43.65% |
| Male03 | 2.8 | 0% | 12.75% | 6.38% | 0.00% | 4.65% | 2.33% |
| Male04 | 2.3 | 0% | 22.29% | 11.15% | 0.00% | 13.91% | 6.96% |
| Male05 | 2.4 | 20% | 20.18% | 20.09% | 20.00% | 7.54% | 13.77% |
| Male06 | 3.4 | 20% | 35.32% | 27.66% | 20.00% | 17.82% | 18.91% |
| Male07 | 3.0 | 0% | 26.50% | 13.25% | 0.00% | 0.30% | 0.15% |
| Male08 | 2.8 | 40% | 9.40% | 24.70% | 40.00% | 0.00% | 20.00% |
| Male09 | 2.7 | 0% | 2.18% | 1.09% | 0.00% | 1.00% | 0.50% |
| Male10 | 1.9 | 0% | 2.22% | 1.11% | 0.00% | 0.00% | 0.00% |
| | Average | 18% | 18.92% | 18.46% | 18.0% | 9.1% | 13.6% |
| | **Overall average** | 15% | 26.03% | 20.51% | **15.0%** | **13.5%** | **14.2%** |

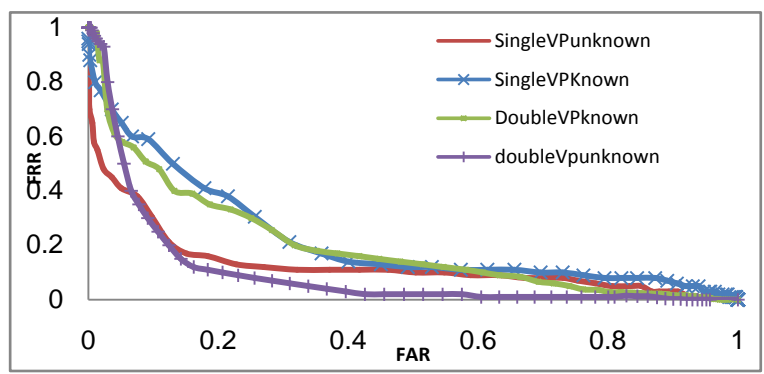

Fig.7. Visual passwords system average error rates during the training stage adding the Chi-square test feature. FRR is a function of FAR.

It can be noticed from figure 7, Table 3 and 4 that the performance of the system has improved by using more than one word in the training set. The overall average error rate is dropped from 15.0% to 8.9% when the password is unknown and from 21.51% to 17.7% when the password is known by the impostors. Even when the password is known, the reported results (of these experiments) emphasis the efficiency of the proposed VP scheme; if the string password is known, in the conventional security systems it is guaranteed by 100% to log into the system. While in the proposed VP scheme even if the password is known still the probability to log into the system is less than 21.51% (depending on the strength of the VP). In addition, the accuracy of the system increased by more than 1% when using extra information such as the chi-square test.

Despite speakers each having their own unique ways of speaking (visual and audio), they still have something in common (e.g. similar phonemes produce similar visemes) – allowing some people to defraud/trick the system to some extent (see Tables 1, 2, 3 and 4). So, the more the client has a matchless, individual way of speaking visually and is consistent, the less chance there is of impostors defrauding his/her visual password (such as female03 and male03, and in contrast with female07 and male06).

Male06 has other problems related to VSR, which include and not limited to the appearance of facial hair. Another problem that affects VSR systems, and therefore the proposed visual passwords system, is the visual speechless persons, who cannot produce a clear visual signal while speaking [3].

Table 3. Visual passwords system results using combination of two Vwords as a VP, "fourfive=>fourfive" in "DoubleVPknown", and "sixseven=>fourfive" in "DoubleVPunknown".

| Subject | Threshold | DoubleVPknown | | | DoubleVPunknown | | |
|---|---|---|---|---|---|---|---|
| | | FRR | FAR | AER | FRR | FAR | AER |
| Female01 | 3.8 | 0% | 29.42% | 14.71% | 0% | 6% | 3% |
| Female02 | 6.1 | 28% | 60.08% | 44.04% | 8% | 27% | 17% |
| Female03 | 3.9 | 0% | 0.42% | 0.21% | 0% | 0.42% | 0.21% |
| Female04 | 4.1 | 0% | 9.89% | 4.95% | 4% | 3.78% | 3.89% |
| Female05 | 6 | 40% | 58.74% | 49.37% | 4% | 1.46% | 2.73% |
| Female06 | 3.6 | 2% | 0.21% | 1.11% | 0% | 1.00% | 0.50% |
| Female07 | 6.4 | 4% | 71.40% | 37.70% | 4% | 84.78% | 44.39% |
| Female08 | 4.4 | 0% | 20.37% | 10.19% | 0% | 4.54% | 2.27% |
| Female09 | 4.6 | 0% | 23.46% | 11.73% | 0% | 2.15% | 1.08% |
| Female10 | 3.6 | 3% | 6.74% | 4.76% | 0% | 0.26% | 0.13% |
| Average | | 7.68% | 28.07% | 17.88% | 2.0% | 13.1% | 7.6% |
| Male01 | 5.1 | 12% | 6.79% | 9.40% | 4% | 0.68% | 2.34% |
| Male02 | 5.1 | 0% | 20.99% | 10.50% | 0% | 0.42% | 0.21% |
| Male03 | 3.9 | 0% | 19.14% | 9.57% | 0% | 0.31% | 0.16% |
| Male04 | 3.8 | 12% | 6.58% | 9.29% | 0% | 1.28% | 0.64% |
| Male05 | 5 | 0% | 20.99% | 10.50% | 28% | 6.69% | 17.35% |
| Male06 | 4.6 | 64% | 37.65% | 50.83% | 44% | 16.94% | 30.47% |
| Male07 | 4.2 | 36% | 11.52% | 23.76% | 24% | 9.90% | 16.95% |
| Male08 | 6 | 8% | 33.74% | 20.87% | 8% | 27.01% | 17.51% |
| Male09 | 3.5 | 32% | 5.35% | 18.68% | 20% | 0.85% | 10.43% |
| Male10 | 3.8 | 12% | 10.70% | 11.35% | 12% | 0.48% | 6.24% |
| Average | | 17.6% | 17.3% | 17.5% | 14.0% | 6.5% | 10.2% |
| **Overall average** | | 12.6% | 22.7% | 17.7% | 8.0% | 9.8% | 8.9% |

Relatively very little work has been reported in the literature concerning the speaker authentication problem using the visual signal alone. One example is the work of Shiell and co-workers [23], their system was evaluated using the "VALID" database, and the reported speaker identification rate was 59.3%. Since the proposed system uses a different database, with respect to system performance, we cannot claim that our system outperform the Shiell's system.

Results in table 5 show the importance and the power of the audio speech when added to the video speech. No doubt that the audio signal is very important when it comes to speech, but there are some problems such as background noise, particularly when providing such passwords in public noisy places, and the willing to reveal such a password loudly in public.

Table 4. Visual passwords system's results, using combination of two Vwords as a VP, "fourfive=>fourfive" in "DoubleVPknown", and "sixseven=>fourfive" in "DoubleVPunknown", adding Chi-square test feature.

| Subject | Threshold | DoubleVPknown | | | DoubleVPunknown | | |
|---|---|---|---|---|---|---|---|
| | | FRR | FAR | AER | FRR | FAR | AER |
| Female01 | 4.3 | 0.00% | 25.38% | 12.69% | 0.00% | 5.00% | 2.50% |
| Female02 | 7.1 | 24.00% | 57.14% | 40.57% | 4.00% | 27.65% | 15.83% |
| Female03 | 4.7 | 0.00% | 1.20% | 0.60% | 0.00% | 1.20% | 0.60% |
| Female04 | 4.7 | 0.00% | 8.00% | 4.00% | 2.00% | 2.00% | 2.00% |
| Female05 | 6.8 | 36.00% | 55.70% | 45.85% | 4.00% | 0.94% | 2.47% |
| Female06 | 4.3 | 2.00% | 0.98% | 1.49% | 0.00% | 2.09% | 1.05% |
| Female07 | 6.8 | 4.00% | 67.59% | 35.80% | 4.00% | 74.54% | 39.27% |
| Female08 | 5.3 | 2.00% | 16.62% | 9.31% | 0.00% | 3.25% | 1.63% |
| Female09 | 5.8 | 0.00% | 20.36% | 10.18% | 0.00% | 0.37% | 0.19% |
| Female10 | 4.1 | 2.00% | 4.50% | 3.25% | 0.00% | 0.10% | 0.05% |
| | Average | 7.00% | 25.75% | 16.37% | 1.40% | 11.71% | 6.56% |
| Male01 | 6.3 | 8.00% | 3.16% | 5.58% | 2.00% | 0.35% | 1.18% |
| Male02 | 6.2 | 0.00% | 16.96% | 8.48% | 0.00% | 0.20% | 0.10% |
| Male03 | 4.7 | 0.00% | 15.47% | 7.74% | 2.00% | 0.60% | 1.30% |
| Male04 | 5.0 | 12.00% | 4.70% | 8.35% | 0.00% | 3.00% | 1.50% |
| Male05 | 5.5 | 0.00% | 18.30% | 9.15% | 24.00% | 7.70% | 15.85% |
| Male06 | 5.7 | 56.00% | 33.50% | 44.75% | 38.00% | 11.54% | 24.77% |
| Male07 | 4.7 | 28.00% | 7.26% | 17.63% | 24.00% | 7.18% | 15.59% |
| Male08 | 6.3 | 8.00% | 30.35% | 19.18% | 8.00% | 24.00% | 16.00% |
| Male09 | 4.2 | 28.00% | 2.00% | 15.00% | 12.00% | 1.33% | 6.67% |
| Male10 | 5.1 | 8.00% | 8.40% | 8.20% | 8.00% | 0.10% | 4.05% |
| | Average | 14.80% | 14.01% | 14.41% | 11.80% | 5.60% | 8.70% |
| | Overall average | 10.90% | 19.88% | 15.39% | 6.60% | 8.66% | 7.63% |

Table 5. Speaker authentication results, accuracy comparison of some methods.

| Work | Audio | Video | Audio/Video |
|---|---|---|---|
| Chibelushi and co-workers [11] | 93.75% | 87.5% | 98.75% |
| Civanlar and Chen [12] | 60% noisy speech<br>95% clean speech | - | 90% noisy speech<br>95% clean speech |
| Chaudhari and co-workers [13] | 95.5% | - | 99.5% |
| Faraj and Bigun [14] | - | - | 98% |
| Shiell and co-workers [17] | - | 59.3% | - |
| Cetingul and co-workers [21] | - | 91.1% - 91.6%<br>88.8% - 91.6% | - |
| Shafait [24] | - | 83% GMM<br>86% KNN | - |
| Hassanat [3] | - | 72% single VP<br>92% double VP<br>small database | - |
| **The proposed work** | - | 79.5%-SingleVPknown<br>85.8%-SingleVPunknown<br>84.6%-DoubleVPknown<br>92.4%-DoubleVPunknown | - |

Whispering the visual password is desirable, since no one want people around to hear their password. From the previous experiments, it is not clear whether the resultant lip motions will be the same as in the case where the sound is revealed. There is a feeling that the air blow may have an alteration on the shape of the lips. Therefore, another experiment is conducted to evaluate the system when uttering the visual passwords without the sound.

This experiment conducted using 15 male students from the IT department in Mutah University, each participant was asked to utter his first name 5 times in 2 different sessions. Students' names were different from each others. The "SingleVPunknown" experiment is used on those data, session 1 is used for training thresholds, and session 2 for testing (see table 6 for the results).

It can be noticed from table 6 (the normal column) that the results of applying the system on a second database support the results obtained from the first database in table 1, and 2. Also it is clear to see the affect of the blown air on the accuracy, the louder the word is spoken, the more blown air, which enhances the shape of lips, and make it better for recognition, and vice versa. When no voice is provided the user need to be a good actor to produce a good visual signal, this is the case of subject 5. In such work personal ability to provide signal is a major concern.

Table 6. Average error rates resulted from experiment "SingleVPunknown" on the second database, adding Chi-square feature.

| subject | vpassword | Threshold | Normal | Whisper | Mute |
|---|---|---|---|---|---|
| 1 | Ibrahim | 4.1 | 23% | 49% | 92% |
| 2 | Audai | 3.6 | 35% | 59% | 84% |
| 3 | Nahar | 2.8 | 12% | 26% | 31% |
| 4 | Firas | 2.3 | 30% | 35% | 42% |
| 5 | Mohammad | 5.3 | 10% | 12% | 8% |
| 6 | AbdelRuhman | 6.1 | 1% | 19% | 35% |
| 7 | Ahmad | 3.1 | 20% | 60% | 81% |
| 8 | Hussam | 4.7 | 19% | 49% | 71% |
| 9 | Mutasim | 3.9 | 8% | 22% | 19% |
| 10 | Omar | 1.9 | 13% | 30% | 61% |
| 11 | Osamah | 3.4 | 12% | 29% | 32% |
| 12 | Ali | 1.7 | 17% | 33% | 48% |
| 13 | Loai | 2.2 | 43% | 59% | 65% |
| 14 | Thamir | 3.6 | 9% | 21% | 28% |
| 15 | Abubakir | 5.1 | 3% | 15% | 18% |
| | Average | | 17% | 34% | 49% |

Generally, if a user wants to provide their visual password mutely, it is their responsibility to do it right and consistent, in this context it is not necessary to say a real word, as the user needs to move their lips in a specific way, and to memorize this way to be used to login next time, but this seems to be a hard task for most people.

**Time complexity analysis for the proposed system**

The total consumed time includes the different stages of the system which are:

Finding the face consumes $O(n^2)$ in one frame; where n is the width and height of each frame. For the whole word it consumes $O(Ln^2)$, where L is the number of frames for each word. Lip localization also consumes $O(Ln^2)$ for each word.

Feature extraction methods has different time complexity depending on each feature, those methods work only on the ROI, which is the mouth area, the size of this area is very small comparing to the size of the face, therefore we assume the mouth width and height equal to a constant k.

Finding the height and width of ROI consumes $O(2Lk)$ for both. Computing M costs $O(Lk^2)$. Computing Q costs $O(Lk^2)$.

Computing R costs $O(Lk^2)$ for wavelet coefficients, $O(Lk^2/4)$ for HL sub band, and, $O(Lk^2/4)$ for LH sub band.

Computing ER costs $O(9Lk^2)$ for vertical edges, and $O(9Lk^2)$ for the horizontal edges, where *9* is the filter size. Computing RC costs $O(Lk^2)$. Computing T costs $O(Lk^2)$. Computing Chi-square costs $O(256Lk^2)$.

Computing KNN costs $O(9Lm)$, where *9* is the number of features, and *m* is the number of samples. Accordingly, the system total time complexity is defined by:

$$T(n)= O(Ln^2) + O(Ln^2) + O(2Lk) + 5\, O(Lk^2) + 2\, O(Lk^2/4) + 2\, O(9Lk^2) + O(256Lk^2) + O(9Lm) \quad (11)$$

The number of frames L is in the range of 30 to 60 frames per word. So it can be considered as a constant number. The number of samples depends on the number of subjects in the system so it must be a variable, and it can be approximated to *n*, accordingly, equation (11) can be approximated to:

$$T(n) = O(kn^2) + O(km) \qquad (12)$$

As a result, time complexity will be quadratic if the size of the frame is greater than the number of samples, and it will be linear if the dimensions of the frame are both small constant numbers. This is the case when motion detection is used, and then the mouth area is the only used part of the frame.

**5. Conclusion and future work**

The visual passwords scheme, which can be thought of as a speaker authentication system using the visual signal alone, was proposed to increase security and reduce identity theft. It is shown in this paper that people with different appearance and unique way of speech visually can produce their passwords using their lips even without having the need for the audio signal, which cannot be produced in some circumstances (acoustic noise). Several experiments were conducted to test the visual passwords system, attaining excellent performance even when the password is known by the imposters. The performance of the proposed visual passwords system can be improved by considering passwords consisting of more than one word. This increases the length of the visual password, and provides a stronger signal, which reduces the probability of being hacked. And therefore, has potential to be a practical approach.

Implementing this work alone isn't provably stronger than many other types of authentication (biometrics), and is just as subject to spoofing/replay attacks as many other forms of authentication. To strengthen the overall authentication system and to avoid replay attacks, the proposed system may make use of session tokens, time stamping or ask the user to provide a part of the visual passwords. For example, the server asks the user to provide the first and the third visual passwords, so a hacker will not get the whole password, and each time the server ask for a different sequence of the visual passwords.

Iris recognition is more accurate than the proposed method, but it needs special equipments. Face recognition might be more accurate as well, but it is easy to trick the system using a picture of the client, audio speech and audio/video speech recognition is also better, but it is affected by the background noise.

The performance of the visual passwords system might be further improved by combining other biometric factors, such as face authentication or even combining or fusing a PIN number, or the string password with the visual word signature. The fusion can be done by concatenating the feature vectors, score-based fusion, or as a decision-based fusing. All these improvements will be investigated in future work.

**6. Acknowledgment**

The author would like to thank professor Sabah Jassim from the University of Buckingham for all his valuable comments and discussions.